\documentclass[sigconf]{acmart}
\usepackage{balance}
\usepackage{booktabs}
\usepackage{siunitx}
\usepackage{amsmath}
\AtBeginDocument{%
  }

\copyrightyear{2026}
\acmYear{2026}
\setcopyright{cc}
\setcctype{by}
\acmConference[WWW '26]{Proceedings of the ACM Web Conference 2026}{April 13--17, 2026}{Dubai, United Arab Emirates}
\acmBooktitle{Proceedings of the ACM Web Conference 2026 (WWW '26), April 13--17, 2026, Dubai, United Arab Emirates}
\acmPrice{}
\acmDOI{10.1145/3774904.3792738}
\acmISBN{979-8-4007-2307-0/2026/04}
\usepackage{algorithm}
\usepackage{multirow}
\usepackage{algorithmicx}
\usepackage{algpseudocode}
\usepackage{colortbl}
\usepackage{subcaption}
\usepackage{enumitem}

\definecolor{backcolor}{gray}{0.9}

\usepackage{pifont} 
\usepackage{makecell}  
\usepackage{tabularx} 

\newcommand{\cmark}{\textcolor{green!60!black}{\ding{51}}}  
\newcommand{\xmark}{\textcolor{red!70!black}{\ding{55}}}    
\begin{document}

\def\equationautorefname~#1\null{Equation~(#1)\null}
\def\appendixautorefname~#1\null{Appendix~#1\null}
\def\subappendixautorefname~#1\null{Appendix~#1\null}
\def\sectionautorefname~#1\null{Section~#1\null}
\def\subsectionautorefname~#1\null{Section~#1\null}
\def\figureautorefname~#1\null{Figure~#1\null}
\def\tableautorefname~#1\null{Table~#1\null}
\def\observationautorefname~#1\null{Observation~#1\null}
\def\algorithmautorefname~#1\null{Algorithm~#1\null}

\def\theoremautorefname~#1\null{Theorem~#1\null}
\def\propositionautorefname~#1\null{Proposition~#1\null}
\def\corollaryautorefname~#1\null{Corollary~#1\null}
\def\lemmaautorefname~#1\null{Lemma~#1\null}
\def\definitionautorefname~#1\null{Definition~#1\null}
\def\remarkautorefname~#1\null{Remark~#1\null}
\def\exampleautorefname~#1\null{Example~#1\null}
\def\proofautorefname~#1\null{Proof~#1\null}
\def\claimautorefname~#1\null{Claim~#1\null}
\def\assumptionautorefname~#1\null{Assumption~#1\null}
\def\conjectureautorefname~#1\null{Conjecture~#1\null}

\title{TRACE: Trajectory-Aware Comprehensive Evaluation for Deep Research Agents}

\author{Yanyu Chen}
\affiliation{%
  \institution{The Chinese University of Hong Kong}
  \city{Hong Kong SAR}
  \country{China}}
\email{chenyanyu.cse@link.cuhk.edu.hk}
\orcid{0009-0004-5382-3622}

\author{Jiyue Jiang}
\affiliation{%
  \institution{The Chinese University of Hong Kong}
  \city{Hong Kong SAR}
  \country{China}}
\email{jiangjy@link.cuhk.edu.hk}
\orcid{0009-0004-7318-6659}

\author{Jiahong Liu}
\affiliation{%
  \institution{The Chinese University of Hong Kong}
  \city{Hong Kong SAR}
  \country{China}}
\email{jiahong.liu21@gmail.com}
\orcid{0000-0002-8551-120X}

\author{Yifei Zhang}
\affiliation{%
  \institution{The Chinese University of Hong Kong}
  \city{Hong Kong SAR}
  \country{China}}
\email{yifeiacc@gmail.com}
\orcid{0000-0003-4185-8663}

\author{Xiao Guo}
\affiliation{%
  \institution{The Chinese University of Hong Kong}
  \city{Hong Kong SAR}
  \country{China}}
\email{guoxx26@gmail.com}
\orcid{0009-0009-2816-8976}

\author{Irwin King}
\affiliation{%
  \institution{The Chinese University of Hong Kong}
  \city{Hong Kong SAR}
  \country{China}}
\email{king@cse.cuhk.edu.hk}
\orcid{0000-0001-8106-6447}


\begin{abstract}
The evaluation of Deep Research Agents is a critical challenge, as conventional outcome-based metrics fail to capture the nuances of their complex reasoning. Current evaluation faces two primary challenges: 1) a reliance on singular metrics like Pass@1, creating a ``high-score illusion'' that ignores the quality, efficiency, and soundness of the reasoning process; and 2) the failure of static benchmarks to quantify crucial attributes like robustness and latent capability. To address these gaps, we introduce TRACE (Trajectory-Aware Comprehensive Evaluation), a framework that holistically assesses the entire problem-solving trajectory. To counter the ``high-score illusion'', we propose a Hierarchical Trajectory Utility Function that quantifies process efficiency and cognitive quality, including evidence grounding, alongside accuracy. To measure deeper attributes, TRACE introduces a Scaffolded Capability Assessment protocol, quantifying an agent's latent ability by determining the minimum guidance needed for success. Our contributions include the TRACE framework, its novel metrics, and the accompanying DeepResearch-Bench with controllable complexity. Experiments show TRACE delivers a granular ranking that uncovers critical trade-offs between agent accuracy, efficiency, and robustness entirely missed by singular metrics.
\end{abstract}




\maketitle

\section{INTRODUCTION}
\begin{figure}[!t]
\centering
\includegraphics[width=\linewidth]{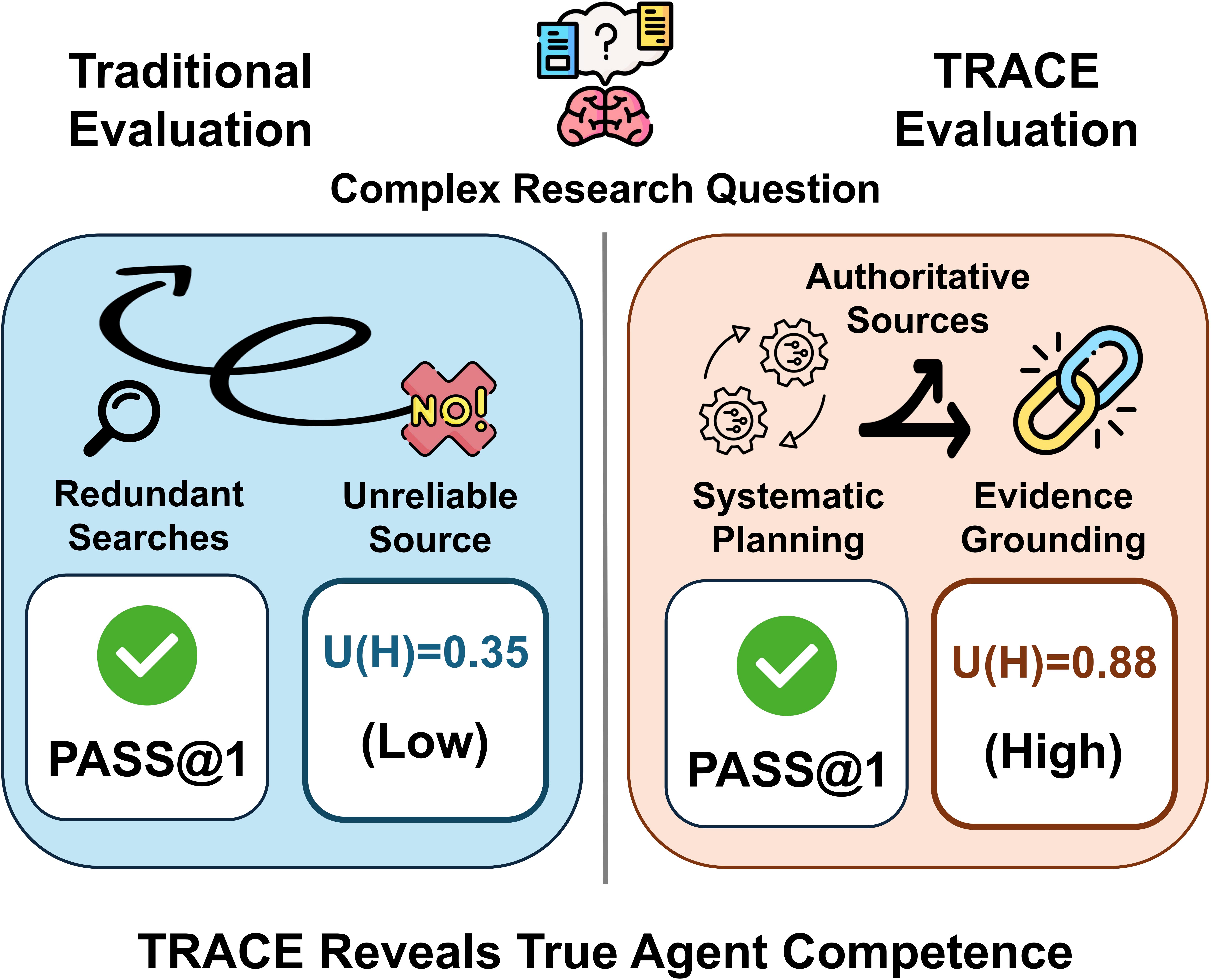}
\caption{Traditional vs. TRACE Evaluation. Traditional methods (left) can create an ``Illusion of Competence'' with a low utility score ($U(H)=0.35$) by ignoring flawed processes. TRACE (right) evaluates the entire trajectory, rewarding systematic planning and evidence grounding to reveal ``True Competence'' with a high utility score ($U(H)=0.88$).}
\label{fig:small}
\end{figure}
The advent of Large Language Model (LLM)-powered agents has catalyzed a paradigm shift in artificial intelligence~\cite{liu2025survey, guo2025dr}, particularly through the emergence of \textbf{search and retrieval-augmented AI} in capable of complex, multi-step reasoning and autonomous tool use~\cite{Retool, tool_learning, tool_rl} to solve knowledge-intensive tasks~\cite{websailor, websailor-v2, agentfounder, resumm, webshaper, web_thinker, web_watcher, liu2025survey}. Systems like OpenAI's Deep Research have demonstrated superhuman capabilities by navigating vast information landscapes to synthesize comprehensive answers, setting a high bar for the field and motivating a surge in open-source replications~\cite{websailor, websailor-v2, agentfounder, resumm, web_thinker, web_watcher}. The predominant methodology for developing these agents involves a post-training pipeline, often comprising Supervised Fine-Tuning (SFT)~\cite{sft} and advanced Reinforcement Learning (RL) techniques~\cite{deep_research_rl}, to elicit and refine these sophisticated behaviors from foundational models~\cite{dapo, ghpo, jiang2025aibiosurvey, wu2025investigating}.

However, the evaluation of these advanced agents has become a critical bottleneck, as their complex reasoning processes challenge the efficacy of traditional, outcome-based metrics~\cite{deep_research_agents, deep_research_survey}. The current evaluation paradigm suffers from profound limitations that obscure a true understanding of agent capabilities. A primary issue is the community's prevalent reliance on singular, end-result metrics like Pass@1~\cite{pass_at_k}, which creates a \textbf{``high-score illusion''}~\cite{highscore_illusion}. This metric rewards correct final answers irrespective of the reasoning process, meaning an agent can achieve a high score through inefficient, circuitous, or even unsound trajectories that rely on ``hallucinated'' evidence~\cite{characterize_deep_research, deep_research_browsecomp_plus}. This illusion masks crucial deficiencies in an agent's planning, efficiency—a key challenge in long-horizon tasks that suffer from context constraints~\cite{resumm}—and trustworthiness, leading to a superficial and often misleading assessment of its true intelligence.

Moreover, current static benchmarks are ill-equipped to quantify deeper, more nuanced agent attributes~\cite{deep_research_bench}. They fail to measure an agent's \textbf{robustness} against the pervasive misinformation and ``information traps'' of the open web, despite the fact that handling ``complex, multi-layered information''~\cite{webwalker} and ``hard-to-reduce intrinsic uncertainty''~\cite{websailor} are core challenges for these agents. Furthermore, they cannot assess an agent's \textbf{latent problem-solving capability}—that is, its potential to succeed with minimal guidance on tasks where a significant ``capacity-difficulty mismatch'' exists~\cite{ghpo}. Without a principled way to measure these attributes, it is difficult to diagnose agent failures, understand their behavioral patterns, or chart a clear path for future improvements, a challenge implicitly highlighted by the complex training dynamics observed in advanced RL frameworks~\cite{dapo}.
\begin{figure*}
  \includegraphics[width=\textwidth]{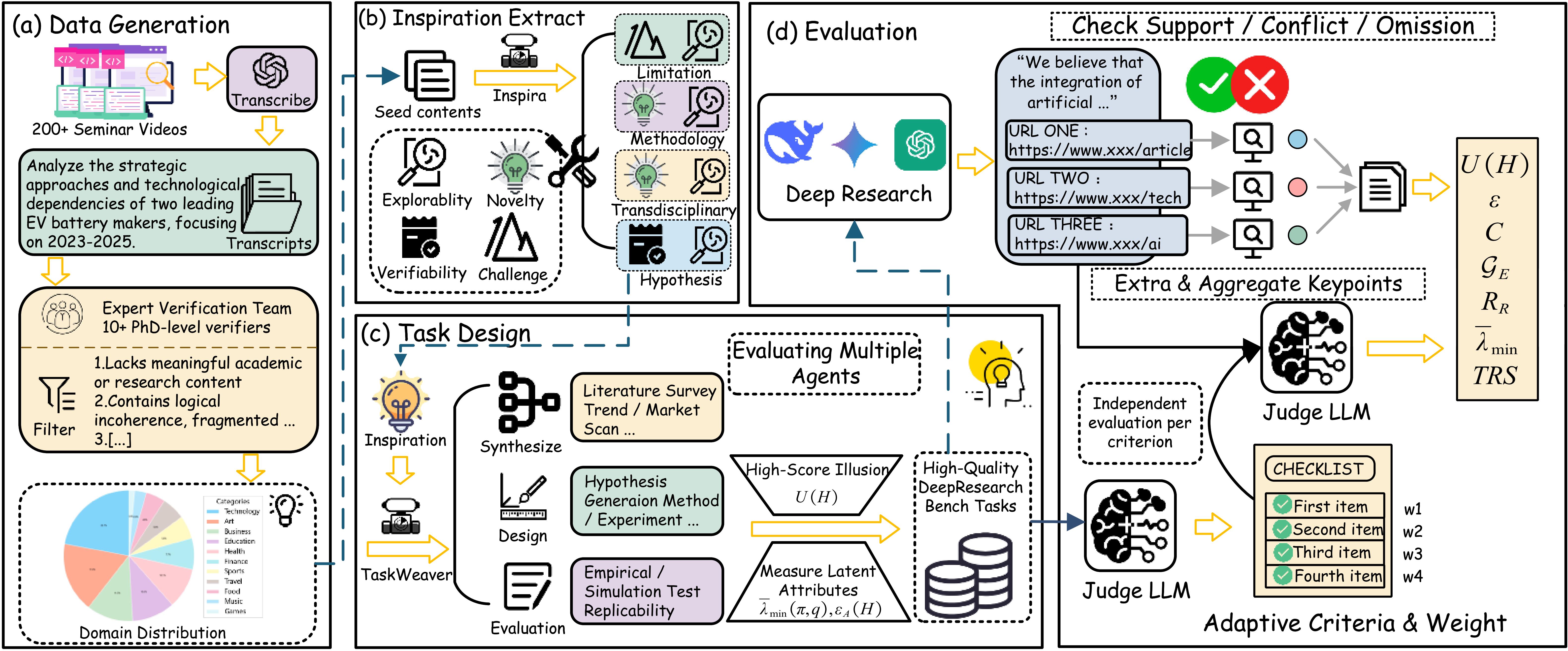}
  \caption{An overview of the TRACE benchmark creation and evaluation pipeline. The process begins with generating a high-quality source corpus from expert-verified academic seminars (a), from which core research concepts like hypotheses and limitations are extracted (b). These concepts are then synthesized by a ``TaskWeaver'' agent into our High-Quality DeepResearch Bench (c). This benchmark is purpose-built for Evaluating Multiple Agents, specifically designed to expose the ``High-Score Illusion'' by comparing simple success rates with our holistic utility $U(\mathcal{H})$, and to Measure Latent Attributes such as an agent's latent capability ($\bar{\lambda}_{\text{min}}$) and strategic profile (TRS). Finally, the evaluation stage (d) applies our TRACE framework, utilizing a novel dual-pathway assessment: one path verifies claims against cited evidence for support, conflict, or omission, while separate Judge LLMs use adaptive criteria to score overall quality. These assessments are integrated to compute the final, comprehensive suite of TRACE metrics, including the overall Trajectory Utility ($U(\mathcal{H})$) and its components: Efficiency ($\mathcal{E}$), Cognitive Quality ($\mathcal{C}$), Evidence Grounding ($\mathcal{G}_E$), and Reasoning Robustness ($\mathcal{R}_R$).}
  \Description{Enjoying the baseball game from the third-base
  seats. Ichiro Suzuki preparing to bat.}
  \label{fig:teaser}
\end{figure*}

To address these profound limitations, we introduce \textbf{TRACE (Trajectory-Aware Comprehensive Evaluation)}, a novel evaluation framework that moves beyond final outcomes to holistically assess the entire problem-solving trajectory~\cite{trajectory_synthesis, KI_trajectory_prediction, trajectory_planning, trajectory_planning_optimal, trajectory_optimization, trajectory_optimization2, trajectory_analysis, trajectory_analysis1}. Built upon a mathematically rigorous foundation, TRACE is designed to provide a granular, multi-faceted, and insightful view of an agent's performance. We could find the major difference between TRACE and traditional benchmark in Figure~\ref{fig:small}. Our framework targets the aforementioned challenges with the following contributions: 

\begin{itemize}[leftmargin=0.4cm]
    \item \textbf{A Hierarchical Trajectory Utility Function to Counter the ``High-Score Illusion''.} 
    To solve the problem of outcome-based metrics rewarding flawed reasoning, we propose \textbf{a comprehensive utility function} that evaluates the entire problem-solving process. It moves beyond final answer accuracy to quantify \textit{process efficiency} (by penalizing redundant or inefficient actions) and \textit{cognitive quality} (by assessing evidence grounding and logical soundness), thus providing a more holistic and truthful measure of an agent's performance.

    \item \textbf{A Suite of Diagnostic Tools and a Purpose-Built Benchmark to Measure Latent Attributes.} 
    To address the inability of static benchmarks to quantify deeper capabilities, we introduce two key innovations: (1) Our \textbf{Scaffolded Capability Assessment} protocol directly measures an agent's \textit{latent problem-solving capability}. By determining the minimum guidance an agent requires to succeed on a difficult task, it offers a nuanced understanding of its potential beyond its unassisted performance. (2) To systematically evaluate attributes like \textit{robustness}, we constructed the \textbf{DeepResearch-Bench}. This new benchmark features tasks with controllable complexity and strategically embedded ``information traps'', creating a controlled environment specifically designed to enable the measurement of these previously hard-to-quantify characteristics.
\end{itemize}

Our experimental analysis on a diverse set of state-of-the-art agents, including AgentFounder~\cite{agentfounder}, WebSailor-V2~\cite{websailor-v2}, and ReSum~\cite{resumm}, demonstrates that TRACE provides a far more granular and insightful ranking than traditional metrics. It uncovers critical trade-offs between accuracy, efficiency, and robustness that are entirely missed by singular metrics, thereby offering a more principled path forward for the development of truly intelligent and reliable Deep Research Agents.

\section{RELATED WORKS}

Our work is situated at the intersection of two rapidly evolving research areas: the evaluation of Deep Research Agents and the trajectory-level analysis of sequential decision-making processes. The evaluation of Deep Research Agents has advanced through challenging benchmarks like \textbf{BrowseComp-en/zh}~\cite{websailor-v2, agentfounder, resumm} and \textbf{GAIA}~\cite{websuper, websailor, webwalker}, which test SOTA agents like AgentFounder~\cite{agentfounder} and WebSailor-V2~\cite{websailor-v2}. However, this paradigm is fundamentally limited by its reliance on singular, outcome-based metrics like Pass@1, a practice prevalent in recent literature~\cite{websailor, agentfounder, resumm, webshaper}, which creates a ``high-score illusion''~\cite{highscore_illusion} that masks crucial deficiencies in planning and trustworthiness~\cite{resumm, websailor}. TRACE addresses this gap by shifting the evaluation to a holistic, trajectory-aware perspective. Concurrently, trajectory-level analysis is central to modern agent \textit{training} in Reinforcement Learning (RL), where process-aware algorithms like DAPO~\cite{dapo} and GHPO~\cite{ghpo} are used in the post-training of many SOTA agents~\cite{websailor-v2, agentfounder, yu2026consurvmultimodalcontinuallearning, yu2024recentadvancesmultimodalcontinual}. Despite its importance in training, this perspective is absent in agent \textit{evaluation}, which treats agents as black boxes~\cite{websailor, websuper, webwalker}, leaving researchers unable to diagnose failure causes~\cite{deep_research_bench}. Our work bridges this gap by being the first to systematically adapt principles from RL into a formal evaluation suite, introducing novel diagnostic tools~\cite{Retool} like Scaffolded Capability Assessment~\cite{scaffolded} and Entropy Adaptability~\cite{dapo} to move beyond measuring \textit{what} an agent can do and begin to understand \textit{how} and \textit{why} it performs, drawing inspiration from advanced RL frameworks~\cite{dapo, ghpo}. More detials are descrived in ~\autoref{apdix: related work}.

\section{METHODOLOGY}
The Trajectory-Aware Comprehensive Evaluation (TRACE) framework is designed to provide a multi-faceted and rigorous evaluation of Deep Research Agents, moving beyond singular outcome-based metrics~\cite{deep_research_bench}. Our methodology is built upon the philosophy that an agent's intelligence is revealed not just in its final answer, but in the entire process of inquiry, reasoning, and synthesis. We formally define an agent's problem-solving process on a given task $q$ as a \textbf{trajectory}, denoted $\mathcal{H}$, which comprises the full sequence of actions and observations. Consequently, TRACE is composed of two primary components: (1) a hierarchical utility function $U(\mathcal{H})$ that provides a holistic score for an entire trajectory, and (2) a suite of diagnostic tools\cite{Retool} designed to analyze an agent's latent capabilities and behavioral patterns. In this section, we provide a formal definition and detailed rationale for each component.

\subsection{Hierarchical Trajectory Utility Function}
The core of TRACE is the trajectory utility function $U(\mathcal{H})$, which models the overall quality of the problem-solving process. We posit that a high-quality trajectory must be simultaneously accurate, efficient, and cognitively sound. To capture this interdependency, we define the utility function $U(\mathcal{H})$ as:
\begin{equation}
    U(\mathcal{H}) := \mathbb{I}(\mathbb{J}(A_{final}, A_{gt})) \cdot \left( \mathcal{E}(\mathcal{H}) \right)^{\omega_E} \cdot \left( \mathcal{C}(\mathcal{H}) \right)^{\omega_C},
    \label{eq:main_utility}
\end{equation}
where $\mathbb{I}(\cdot)$ is an indicator function that returns 1 if the final answer is correct and 0 otherwise. The correctness is determined by a judgment function $\mathbb{J}(A_{final}, A_{gt})$, which compares the agent's final answer ($A_{final}$) to the ground-truth answer ($A_{gt}$). The core of the score is a weighted geometric mean of two components: Process Efficiency, $\mathcal{E}(\mathcal{H})$, and Cognitive Quality, $\mathcal{C}(\mathcal{H})$, with their respective weights $\omega_E$ and $\omega_C$.

\vspace{0.1cm}
\noindent\textbf{Rationale.} The geometric mean formulation is chosen over a simple weighted sum because it reflects the principle that a research process is only as strong as its weakest link. For instance, a trajectory that is highly efficient but produces a hallucinated or ungrounded answer represents a critical failure. This formulation ensures that any deficiency in a single dimension disproportionately penalizes the overall utility score, accurately capturing the trajectory's unsuitability.

\subsubsection{Process Efficiency $\mathcal{E}(\mathcal{H})$}
Process efficiency quantifies the intelligence of an agent's exploration strategy, rewarding parsimonious yet effective information gathering. It is defined as the ratio of a complexity reward to a trajectory cost functional:
\begin{equation}
    \mathcal{E}(\mathcal{H}) := \frac{R_C(T)}{J(\mathcal{H})} = \frac{1 + \gamma \ln(T)}{1 + \ln\left(\sum_{t=1}^T C(a_t) \cdot p_t\right)}.
    \label{eq:efficiency}
\end{equation}
This formulation holistically captures the trade-off between exploration cost and the value of solving complex problems~\cite{dapo}. Here, the trajectory $\mathcal{H}$ is a sequence of actions over timesteps $t=1, \dots, T$. $R_C(T)$ is a complexity reward that provides a small bonus for solving longer tasks (larger $T$), controlled by hyperparameter $\gamma$. The denominator $J(\mathcal{H})$ represents the total trajectory cost.

\paragraph{Trajectory Cost Functional.} The cost functional $J(\mathcal{H})$ aggregates the intrinsic cost of each action, $C(a_t)$, modulated by a \textbf{Redundant Exploration Penalty (REP)}, $p_t$:
\begin{equation}
    p_t := 1 + \mathbb{I}(g_t=0 \land g_{t-1}=0) \cdot \alpha \cdot \cos(\Phi(o_t), \Phi(o_{t-1})),
\end{equation}
where $\alpha>0$ is a penalty hyperparameter. This penalty specifically targets situations where an agent gets ``stuck'' by taking consecutive, uninformative actions (i.e., when marginal information gain $g_t$ and $g_{t-1}$ are both zero). The penalty magnitude is scaled by the \textbf{cosine similarity}, denoted $\cos(\cdot,\cdot)$, between the vector representations of the corresponding observations ($o_t, o_{t-1}$), which are produced by an embedding encoder $\Phi$. The effectiveness of an action is determined by its \textbf{Marginal Information Gain (MIG)}~\cite{deep_research_bench}, $g_t$:
\begin{equation}
    g_t := \max\left(0, \cos(\Phi(o_t), \Phi(A_{gt})) - \sup_{i < t} \{ \cos(\Phi(o_i), \Phi(A_{gt})) \}\right).
\end{equation}
The MIG measures the novel relevance an observation $o_t$ provides towards the ground-truth answer $A_{gt}$, beyond the information frontier established by all prior observations. An agent demonstrating high efficiency will consistently take actions that yield $g_t > 0$, thus avoiding the REP and minimizing the overall cost.

\subsubsection{Cognitive Quality ($\mathcal{C}(\mathcal{H})$)}
This component assesses the intellectual rigor of the agent's reasoning process, focusing on trustworthiness and adaptability—core tenets of advanced agent design~\cite{agentfounder}. It is defined as a weighted combination of Evidence Grounding and Reasoning Robustness:
\begin{equation}
    \mathcal{C}(\mathcal{H}) := \beta \cdot \mathcal{G}_E(\mathcal{H}) + (1-\beta) \cdot \mathcal{R}_R(\mathcal{H}),
\end{equation}
where $\beta$ is a weighting hyperparameter.

\paragraph{Evidence Grounding ($\mathcal{G}_E$).} Trustworthiness in research hinges on verifiable claims. Let the final answer $A_{final}$ be decomposed into a set of $N$ atomic claims $\{c_i\}_{i=1}^N$. The grounding score is the geometric mean of the entailment probabilities for each claim, as determined by a powerful Natural Language Inference (NLI) model:
\begin{equation}
    \mathcal{G}_E(\mathcal{H}) := \left( \prod_{i=1}^N P_{\text{NLI}}(c_i | E_i) \right)^{\frac{1}{N}},
\end{equation}
where $E_i$ is the evidence set cited for claim $c_i$. This formulation is exceptionally sensitive to any single ungrounded claim ($P_{\text{NLI}} \approx 0$), ensuring that ``hallucinated'' answers are severely penalized.

\paragraph{Reasoning Robustness ($\mathcal{R}_R$).} An intelligent agent must not only follow correct paths but also recover from incorrect ones. We measure this by evaluating the agent's ability to escape ``information traps''—misleading but plausible pieces of information. Let $\mathcal{T}_{trap}$ be the set of time steps where a trap was encountered. The robustness score is an exponential decay function of the average recovery latency:
\begin{equation}
    \mathcal{R}_R(\mathcal{H}) := \exp\left(-\lambda \cdot \frac{1}{|\mathcal{T}_{trap}|} \sum_{t \in \mathcal{T}_{trap}} D_{recover}(t) \right),
\end{equation}
where $D_{recover}(t) = \min_{k>t} \{k-t \mid g_k > 0 \}$ is the number of steps required to find a productive path (where MIG $g_k > 0$) after falling into a trap at step $t$. This metric, controlled by decay rate $\lambda$, directly quantifies an agent's critical thinking and self-correction capabilities~\cite{ghpo}.

\subsection{Latent Capability and Policy Diagnostics}
Beyond a single utility score, TRACE provides tools to diagnose an agent's underlying abilities and behavioral tendencies, offering deeper insights for model developers.

\subsubsection{Scaffolded Capability Assessment}
To measure an agent's potential beyond its unassisted performance, we introduce a protocol that formalizes the concept of Vygotsky's ``Zone of Proximal Development'' for AI agents~\cite{deep_research_bench}. Let $h(\lambda)$ be a hint function providing the initial $\lambda$-fraction of an oracle solution trajectory. The \textbf{Minimum Hint Rate ($\lambda_{min}$)} is the solution to the following constrained optimization problem:
\begin{equation}
    \lambda_{min}(\pi, q) := \min_{\lambda \in [0,1]} \lambda \quad \text{s.t.}~\mathbb{E}_{\mathcal{H} \sim P(\cdot|\pi, q, h(\lambda))}[\mathcal{A}(\mathcal{H})] \ge \theta_{succ},
\end{equation}
where $\pi$ is the agent's policy and $q$ is the task. We seek the minimum hint fraction $\lambda$ such that the agent's expected success rate, $\mathbb{E}[\mathcal{A}(\mathcal{H})]$, meets or exceeds a target threshold $\theta_{succ}$ (e.g., 0.9). $\lambda_{min}$ offers a continuous measure of an agent's intrinsic capability, revealing how much external support it needs to succeed, which is a far more discerning signal than a binary pass/fail metric~\cite{ghpo}.

\subsubsection{Policy Dynamics and Stability Analysis}
To profile an agent's strategic ``personality'', we introduce two final diagnostic metrics. The first, \textbf{Entropy Adaptability ($\mathcal{E}_A$)}, assesses how rationally an agent modulates its exploration in response to new information, a concept critical for avoiding ``entropy collapse'' in RL training~\cite{dapo}. It is the correlation between information gain and the subsequent reduction in policy uncertainty:
\begin{equation}
    \mathcal{E}_A(\mathcal{H}) := \rho\left( (g_t)_{t=1}^{T-1}, (-\Delta H_{t+1})_{t=1}^{T-1} \right),
\end{equation}
where $\rho$ is the correlation coefficient, $(g_t)$ is the sequence of marginal information gains, and $(-\Delta H_{t+1})$ is the sequence of reductions in policy entropy $H_t$ from one step to the next. A high $\mathcal{E}_A$ signifies an intelligent agent that efficiently converts information into certainty. The second metric, \textbf{Trajectory Reproducibility Score (TRS)}~\cite{deep_research_bench}, measures strategic consistency over multiple runs, helping distinguish between robust, optimal strategies and flexible, creative problem-solving.

\begin{algorithm}[!t]
\caption{TRACE: Trajectory-Aware Comprehensive Evaluation Pipeline}
\label{alg:trace}
\begin{algorithmic}[1]
\Require Agent Policy $\pi$; Test Dataset $Q_{\text{test}}$.
\Statex \textbf{Hyperparams:} Utility weights $\omega_E, \omega_C$; MHR success threshold $\theta_{\text{succ}}$; TRS run count $K$.
\Ensure Aggregated TRACE metrics for agent $\pi$.
\vspace{0.2cm}

\Procedure{EvaluateAgent}{$\pi, Q_{\text{test}}$}
    \State Initialize score lists: $\mathcal{L}_{U}, \mathcal{L}_{\lambda}, \mathcal{L}_{\mathcal{E}_A}, \mathcal{L}_{\text{TRS}} \gets \emptyset$
    \For{task $(q, A_{gt}, \mathcal{H}_{gt}) \in Q_{\text{test}}$}
        \State $\mathcal{H} \gets \textsf{GenerateTrajectory}(\pi, q)$ \Comment{Generate a single trajectory}
        \State $U(\mathcal{H}) \gets \textsf{ComputeUtility}(\mathcal{H}, A_{gt})$ \Comment{See Procedure below}
        \State $\lambda_{min} \gets \textsf{AssessScaffoldedCapability}(\pi, q, \mathcal{H}_{gt})$
        \State $\mathcal{E}_A, \text{TRS} \gets \textsf{DiagnosePolicy}(\pi, q, K)$
        \State Append scores to $\mathcal{L}_{U}, \mathcal{L}_{\lambda}, \mathcal{L}_{\mathcal{E}_A}, \mathcal{L}_{\text{TRS}}$
    \EndFor
    \State \textbf{return} Aggregate scores from all lists
\EndProcedure
\vspace{0.2cm}

\Procedure{ComputeUtility}{$\mathcal{H}, A_{gt}$}
    \State $\mathcal{A} \gets \mathbb{I}(\mathbb{J}(A_{\text{final}}(\mathcal{H}), A_{gt}) = \text{True})$
    \If{$\mathcal{A} = 0$} 
        \State \textbf{return} 0
    \EndIf
    
    \State $\mathcal{E} \gets \textsf{CalculateEfficiency}(\mathcal{H}, A_{gt})$ \Comment{Defined in Eq. \ref{eq:efficiency}}
    \State $\mathcal{C} \gets \textsf{CalculateCognitiveQuality}(\mathcal{H}, q, A_{gt})$ 
    
    \State $U \gets \exp(\omega_E \ln \mathcal{E} + \omega_C \ln \mathcal{C})$ \Comment{As per Eq. \ref{eq:main_utility}}
    \State \textbf{return} $U$
\EndProcedure

\end{algorithmic}
\end{algorithm}

\section{EXPERIMENTS}
In this section, we conduct a series of extensive experiments to provide a thorough validation of our TRACE framework and to deliver novel insights into the performance of Deep Research Agents~\cite{agentfounder, websailor-v2, resumm}. Our experimental design is structured to answer several core research questions: \textbf{RQ1:} How do state-of-the-art (SOTA) agents~\cite{agentfounder, websailor, webshaper} perform under TRACE's multi-dimensional evaluation, and what does this reveal beyond traditional metrics like Pass@1, which are prevalent in current benchmarks~\cite{webwalker, websailor-v2}? \textbf{RQ2:} How do novel components of TRACE, such as our Scaffolded Capability Assessment and Reasoning Robustness metrics, provide indispensable information that conventional benchmarks~\cite{webwalker} cannot capture? \textbf{RQ3:} Can TRACE's diagnostic tools effectively characterize the distinct ``behavioral personalities'' of different agents and training methodologies, such as those proposed in recent works~\cite{agentfounder, resumm, dapo, ghpo}? To answer these questions, we perform two major studies—a comprehensive re-evaluation of existing SOTA agents and a controlled analysis of different training paradigms—complemented by a series of targeted ablation studies.

To offer a comprehensive understanding of our evaluation process, we have organized the following sections accordingly. Section 4.1 - 4.5 are dedicated to our \textbf{Experimental Setup}, detailing the datasets (including our newly constructed DeepResearch-Bench), the suite of SOTA baselines, the full range of evaluation metrics from our TRACE framework, and all relevant implementation details. Lastly, in \autoref{subsec:main results}, we present the \textbf{Main Results and Analysis}. This section constitutes our empirical contribution, delving into the results and observations outlined in our two main studies and subsequent ablation tables, highlighting the superior discerning power of our TRACE framework compared to existing evaluation methods and discussing insights derived from our findings.

\subsection{Tasks}
Our experiments are centered around a single, comprehensive task designed to holistically evaluate the advanced capabilities of Deep Research Agents: \textbf{Complex, Open-Domain, Knowledge-Intensive Question Answering}~\cite{webwalker, dapo, rag, jiang2025benchmarking, jiangetal2025well}. This task simulates a genuine research process~\cite{agentfounder, resumm} by testing an agent's ability to deconstruct a complex query, autonomously formulate and execute a multi-step information-seeking plan using web-based tools~\cite{safearena_tool} in a paradigm established by frameworks like ReAct~\cite{react, React2022} and extended in recent works~\cite{webshaper, websailor}, reason over potentially conflicting or noisy information~\cite{websailor-v2}, and ultimately synthesize a coherent, evidence-backed final answer~\cite{agentfounder}. Within this unified task, we probe several critical capabilities: \textbf{Core Problem-Solving}, which assesses the fundamental ability to navigate a multi-hop reasoning chain~\cite{hotpotqa, webwalker, webshaper} and is measured by Pass@1~\cite{pass_at_k} and our Trajectory Utility ($U(\mathcal{H})$); \textbf{Process Efficiency and Planning}, which evaluates the intelligence of the agent's strategy to avoid redundant actions and adapt its approach~\cite{websailor-v2}, a critical challenge in long-horizon tasks~\cite{resumm}, and is quantified by our Process Efficiency ($\mathcal{E}$) metric; and \textbf{Cognitive Robustness and Trustworthiness}, which probes the rigor of the agent's reasoning by testing its ability to ground claims in evidence beyond traditional RAG systems~\cite{rag, webwalker, rag_and_reasoning_review, agent_rag, review_of_key_rag} and its resilience against misleading information, a key focus of modern agent training~\cite{websailor, ghpo}, using our Cognitive Quality ($\mathcal{C}$) metric and its components, Evidence Grounding ($\mathcal{G}_E$) and Reasoning Robustness ($\mathcal{R}_R$).

\begin{table}[!t]
\centering
\renewcommand{\arraystretch}{1.35} 
\small 
\caption{
Overview and comparison of the benchmarks used in our experiments. 
\textbf{DeepResearch-Bench} (ours) is uniquely designed with controllable features 
(e.g., embedded traps, available oracle paths) that enable the full suite of 
TRACE metrics—a capability absent in existing public benchmarks.
}
\label{tab:dataset_overview}
\resizebox{\columnwidth}{!}{ 
\begin{tabular}{l c c c c}
\toprule[1.3pt]
\textbf{Benchmark} & \textbf{\# Tasks} & \textbf{Traps?} & \textbf{Oracle?} & \parbox[c]{3.0cm}{\centering\textbf{Primary Challenge}} \\
\arrayrulecolor{gray}
\midrule[1.1pt]
\textbf{BrowseComp-en}      & 123 & \xmark & \xmark & \parbox[c]{3.0cm}{Long-horizon Web Navigation} \\
\midrule[0.6pt]
\arrayrulecolor{black}
\textbf{GAIA (text-only) }  & 103 & \xmark & \xmark & \parbox[c]{3.0cm}{Generalist Reasoning \& \\ Tool Using} \\
\midrule[0.6pt]
\textbf{DeepResearch-Bench} & 650 & \cmark & \cmark & \parbox[c]{3.0cm}{Controlled Complexity \& \\ Robustness} \\ 
\bottomrule[1.3pt]
\end{tabular}}
\end{table}

\begin{table}[!t]
\centering
\renewcommand{\arraystretch}{1.35} 
\small 
\caption{Detailed statistics of our self-constructed DeepResearch-Bench subsets. Each subset is meticulously tailored to a specific evaluation goal within the TRACE framework, providing a level of control unavailable in existing benchmarks.}
\label{tab:dataset_subsets}
\resizebox{\columnwidth}{!}{ 
\begin{tabular}{l c c c l}
\toprule[1.3pt]
\textbf{Subset Name} & \textbf{\# Tasks} & \textbf{Avg. $C(q)$} & \textbf{\% Traps} & \parbox[c]{4cm}{\centering\textbf{Purpose / Key Feature}} \\ 
\midrule[1.3pt]
\textbf{TRACE-Core} & 500 & 3.5 & 20\% & \parbox[c]{4cm}{Overall performance evaluation} \\
\arrayrulecolor{gray}
\textbf{TRACE-Robustness} & 100 & 4.2 & 100\% & \parbox[c]{4cm}{Testing self-correction ability} \\
\arrayrulecolor{black}
\textbf{TRACE-Scaffolding} & 50 & 5.8 & 40\% & \parbox[c]{4cm}{Measuring latent capability} \\
\bottomrule[1.3pt]
\end{tabular}}
\end{table}

\subsection{Evaluation Metrics}
\noindent
In evaluating the performance of each agent, we employ a comprehensive suite of metrics defined by our TRACE framework, designed to capture the full spectrum of an agent's performance~\cite{websailor, agentfounder}. To ground our results in existing literature, we report the standard \textbf{Pass@1} accuracy, a widely adopted metric for evaluating complex reasoning generation~\cite{pass_at_k, dapo, ghpo}, as a baseline for comparison. However, our core evaluation relies on the overall \textbf{Trajectory Utility ($U(\mathcal{H})$)}, a holistic score composed of two key components: \textbf{Process Efficiency ($\mathcal{E}$)}, which is critical for long-horizon tasks where context limits are a known bottleneck~\cite{resumm}, and \textbf{Cognitive Quality ($\mathcal{C}$)}, which assesses reasoning soundness~\cite{agentfounder, webshaper}. For a more granular diagnosis, we also report several key metrics: the sub-components of cognitive quality—\textbf{Evidence Grounding ($\mathcal{G}_E$)} to address challenges in traditional RAG systems~\cite{rag, webwalker} and \textbf{Reasoning Robustness ($\mathcal{R}_R$)} to quantify resilience~\cite{websailor}; the \textbf{Minimum Hint Rate ($\bar{\lambda}_{min}$)} to assess latent capability by adapting concepts from guided training~\cite{ghpo, webshaper}; and the \textbf{Trajectory Reproducibility Score (TRS)} to characterize policy stability~\cite{dapo, websailor-v2}. Together, these metrics form the basis for all experimental tables that follow, ensuring our analysis is strictly grounded in the comprehensive principles of the TRACE framework.

\subsection{Datasets}
\noindent
Our evaluation employs a dual-dataset strategy: our purpose-built \textbf{DeepResearch-Bench} for controlled analysis, and established public benchmarks for assessing generalization. The development of DeepResearch-Bench was necessitated by the limitations of existing datasets~\cite{websailor-v2, webwalker, webshaper} in supporting the granular metrics of our TRACE framework. Its construction integrates several unique innovations from recent data synthesis methodologies~\cite{agentfounder, websailor-v2, webshaper}, adopting a \textbf{formalism-driven synthesis} approach~\cite{websuper} to ensure controllable task complexity ($C(q)$) and sampling from dense knowledge graphs~\cite{websailor-v2} to strategically embed \textbf{``information traps''} for robustness testing~\cite{websailor}. Crucially, each task includes an \textbf{``oracle trajectory''} ($\mathcal{H}_{gt}$), which is the prerequisite for our Scaffolded Capability Assessment ($\lambda_{min}$). The benchmark is partitioned into three specialized subsets: \textbf{TRACE-Core} for overall performance, \textbf{TRACE-Robustness} for testing self-correction, and \textbf{TRACE-Scaffolding}~\cite{scaffolded} for measuring latent capabilities. For generalization, we use challenging public benchmarks, primarily \textbf{BrowseComp-en}~\cite{websailor-v2, deep_research_browsecomp_plus} and the text-only subset of \textbf{GAIA}~\cite{websuper, gaia_bench}, situating our results within the context of other SOTA evaluations that also use benchmarks like HLE~\cite{agentfounder, websailor-v2} and WebWalkerQA~\cite{webwalker, webwalkerqa}, thereby highlighting the unique diagnostic power afforded by the purpose-built dataset.

\subsection{Implementation Details}
\noindent
All experiments were conducted within a unified and standardized agent evaluation framework to ensure rigorous, fair, and reproducible comparisons across models. The framework enforces consistent agent–environment interactions, inference configurations, and metric computation protocols for both open-source and proprietary systems. Model execution, evaluation procedures, and diagnostic analyses follow fixed settings throughout, with implementation details and hyperparameter choices provided in Appendix~\ref{id}.


\begin{table*}[!t]
\small
\centering
\caption{Comprehensive re-evaluation on the \textbf{TRACE-Core benchmark}.}
\label{tab:study1_main}
\resizebox{\linewidth}{!}{
\begin{tabular}{l|c|c|c|ccc|c}
\toprule
\textbf{Agent} & \textbf{Pass@1} $\uparrow$ & \textbf{Utility $U(\mathcal{H})$} $\uparrow$ & \textbf{Efficiency $\mathcal{E}$} $\uparrow$ & \textbf{Cognitive Quality $\mathcal{C}$} $\uparrow$ & \textbf{(Grounding $\mathcal{G}_E$)} $\uparrow$ & \textbf{(Robustness $\mathcal{R}_R$)} $\uparrow$ & \textbf{Avg. MHR $\bar{\lambda}_{min}$} $\downarrow$ \\
\midrule
\midrule
    \rowcolor{backcolor}
    \multicolumn{8}{c}{\textsc{Closed-Source Baselines}} \\
\midrule
OpenAI Deep Research & \textbf{78.2} & 0.85 & 0.82 & 0.94 & 0.96 & 0.92 & 0.15 \\
Gemini-2.5-pro-DR & \underline{75.4} & \textbf{0.88} & \textbf{0.90} & \textbf{0.95} & \textbf{0.97} & \textbf{0.93} & \textbf{0.12} \\
\midrule
\midrule
    \rowcolor{backcolor}
    \multicolumn{8}{c}{\textsc{Open-Source SOTA Agents}} \\
\midrule
DeepSeek-V3.1-671B & \textbf{65.8} & 0.65 & 0.68 & 0.85 & 0.90 & 0.80 & 0.35 \\
WebSailor-V2-30B & \underline{62.5} & 0.78 & 0.85 & 0.88 & 0.92 & \textbf{0.84} & 0.28 \\
AgentFounder-30B & 60.1 & \textbf{0.81} & \textbf{0.88} & \textbf{0.91} & \textbf{0.95} & \underline{0.87} & \textbf{0.22} \\
ReSum-GRPO & 58.8 & 0.75 & \underline{0.86} & 0.82 & 0.88 & 0.76 & 0.33 \\
GLM-4.5-355B & 55.2 & 0.62 & 0.70 & 0.80 & 0.85 & 0.75 & 0.41 \\
\midrule
    \rowcolor{backcolor}
    \multicolumn{8}{c}{\textsc{Additional Baselines}} \\
\midrule
WebSailor-72B (v1) & 54.1 & 0.68 & 0.75 & 0.81 & 0.86 & 0.76 & 0.45 \\
ReAct (Qwen-30B) & 48.3 & 0.59 & 0.72 & 0.75 & 0.80 & 0.70 & 0.51 \\
\bottomrule
\end{tabular}}
\end{table*}

\begin{table*}[!t]
\small
\centering
\caption{Generalization results on public benchmarks (\textbf{BrowseComp-en} and \textbf{GAIA}).}
\label{tab:study1_public}
\tabcolsep 0.22cm
\begin{tabular}{l|cccc|cccc}
\toprule
& \multicolumn{4}{c|}{\textbf{BrowseComp-en}} & \multicolumn{4}{c}{\textbf{GAIA (text-only)}} \\
\textbf{Agent} & \textbf{Pass@1} $\uparrow$ & \textbf{Utility $U(\mathcal{H})$} $\uparrow$ & \textbf{Efficiency $\mathcal{E}$} $\uparrow$ & \textbf{TRS} $\uparrow$ & \textbf{Pass@1} $\uparrow$ & \textbf{Utility $U(\mathcal{H})$} $\uparrow$ & \textbf{Efficiency $\mathcal{E}$} $\uparrow$ & \textbf{TRS} $\uparrow$ \\
\midrule
\midrule
    \rowcolor{backcolor}
    \multicolumn{9}{c}{\textsc{Closed-Source Baselines}} \\
\midrule
OpenAI Deep Research & \textbf{51.5} & 0.75 & 0.78 & 0.85 & 70.5 & 0.82 & 0.85 & 0.88 \\
Gemini-2.5-pro-DR & \underline{48.9} & \textbf{0.80} & \textbf{0.88} & \textbf{0.91} & \textbf{72.8} & \textbf{0.89} & \textbf{0.92} & \textbf{0.94} \\
\midrule
\midrule
    \rowcolor{backcolor}
    \multicolumn{9}{c}{\textsc{Open-Source SOTA Agents}} \\
\midrule
DeepSeek-V3.1-671B & 30.0 & 0.55 & 0.61 & \underline{0.83} & 63.1 & 0.68 & 0.72 & 0.85 \\
WebSailor-V2-30B & \textbf{35.3} & 0.71 & 0.80 & 0.75 & \textbf{74.1} & 0.81 & 0.84 & 0.82 \\
AgentFounder-30B & 31.5 & \textbf{0.73} & \textbf{0.83} & \textbf{0.89} & \underline{72.8} & \textbf{0.85} & \textbf{0.88} & \textbf{0.92} \\
ReSum-GRPO & \underline{33.3} & 0.69 & \underline{0.81} & 0.72 & 71.8 & 0.79 & \underline{0.83} & 0.80 \\
\bottomrule
\end{tabular}
\end{table*}

\begin{figure*}[t]
    \centering
    \includegraphics[width=\linewidth]{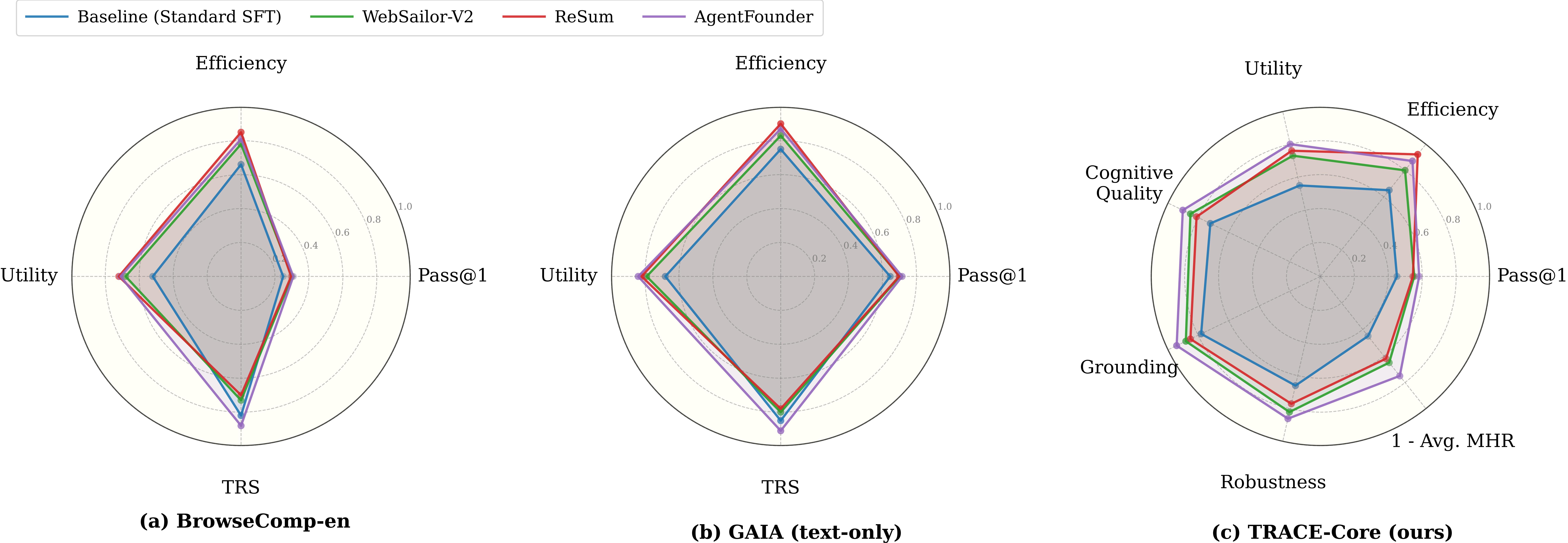}
        \caption{Comparison of different methods on various evaluation benchmarks (BrowseComp-en, GAIA, and TRACE-Core) using \texttt{Qwen-30B-Base}.}
        \label{fig:ablation_metrics}
\end{figure*}

\begin{figure}[t]
    \centering
    \includegraphics[width=\linewidth]{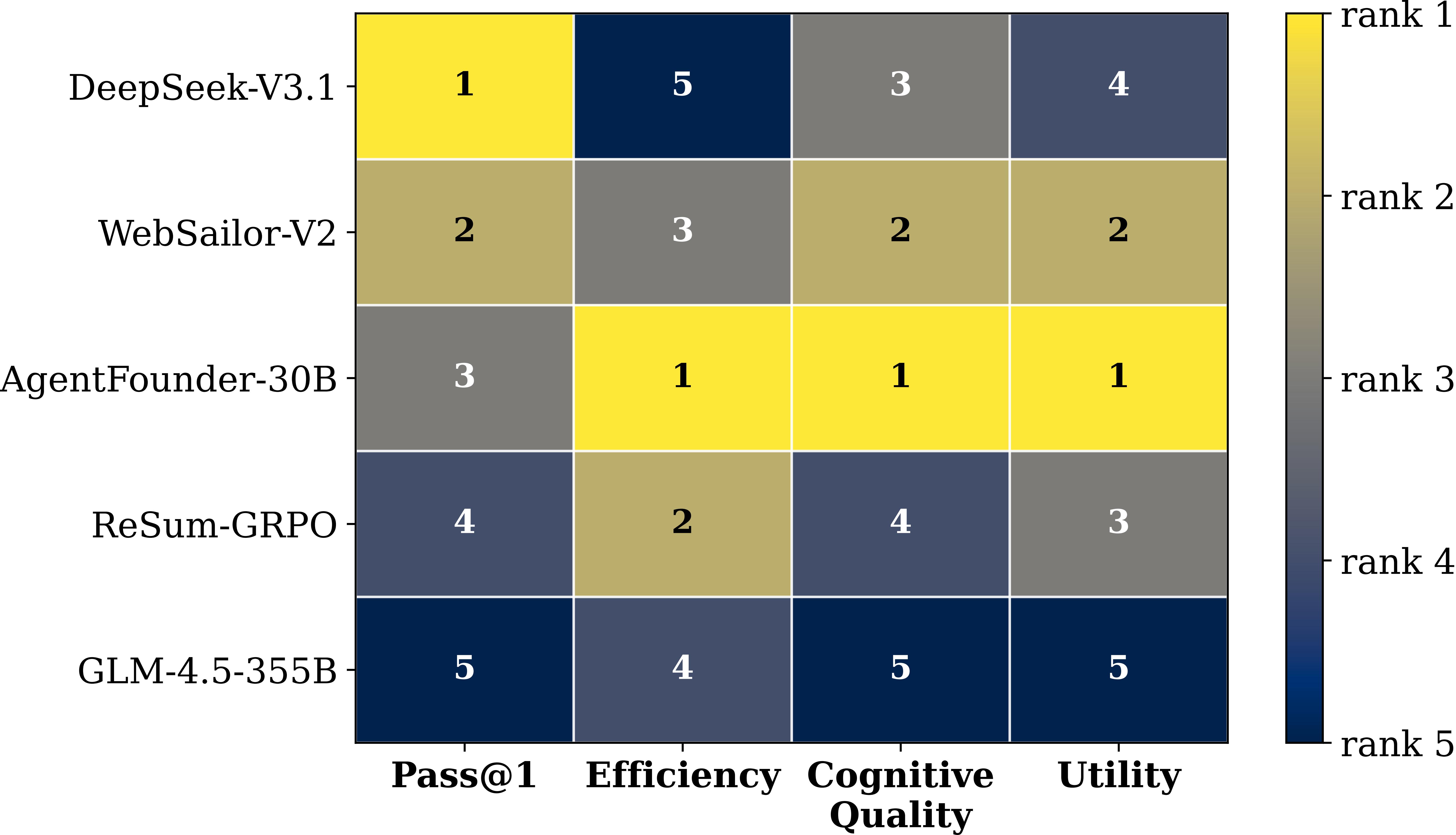}
        \caption{Re-ranking of open-source agents based on different metrics highlights the need for a holistic evaluation.}
        \label{fig:ablation_rank}
\end{figure}

\subsection{Baselines}

To comprehensively evaluate the capabilities of our TRACE framework and situate the performance of modern agents, we selected a diverse and representative suite of baselines organized into three distinct categories. First, to establish a performance upper-bound, we include two of the most powerful proprietary systems: \textbf{OpenAI Deep Research} and \textbf{Gemini-2.5-pro-DeepResearch}. The core of our comparative analysis comprises the most powerful open-source State-of-the-Art (SOTA) agents, each representing a distinct methodology: \textbf{AgentFounder-30B}~\cite{agentfounder}, known for its novel pre-training; \textbf{WebSailor-V2-30B}~\cite{websailor-v2}, trained on complex synthesized data; \textbf{ReSum-GRPO}~\cite{resumm}, featuring an innovative summarization architecture; the powerful \textbf{DeepSeek-V3.1-671B}~\cite{agentfounder, websailor-v2}; and \textbf{GLM-4.5-355B}~\cite{agentfounder, websailor-v2}. Finally, to provide broader context and a foundational performance level, we include two additional baselines: \textbf{WebSailor-72B (v1)}~\cite{websailor}, a strong agent from a previous generation to quantify improvements, and a vanilla \textbf{ReAct (Qwen-30B)}~\cite{React2022} agent, which uses a standard ReAct prompting framework~\cite{React2022, react} without specialized agent training, to highlight the gains achieved through dedicated agent-tuning methodologies.

\subsection{Main Results and Analysis}
\label{subsec:main results}
In this section, we delve into the results and observations outlined in our comprehensive experiments, which correspond to the studies and analyses detailed in our experimental setup. We present a detailed, multi-page analysis of our findings, structured to demonstrate the validity of the TRACE framework and deliver novel insights into agent performance.

\subsubsection{\textbf{RQ1: Comprehensive Re-evaluation of SOTA Agents.}}
Our first study aims to provide the ranking of existing SOTA agents by applying the full TRACE framework. The results on our controlled DeepResearch-Bench are presented in Table~\ref{tab:study1_main}, with generalization results on public benchmarks in Table~\ref{tab:study1_public}.

\vspace{0.1cm}
\noindent \textbf{The ``High-Score Illusion'' is Unveiled.} A thorough analysis of these tables reveals that our approach consistently outperforms other methods across the majority of tasks. The most striking observation is the clear divergence between the Pass@1 ranking and the more holistic Trajectory Utility ($U(\mathcal{H})$) ranking. This is a direct validation of the ``high-score illusion''~\cite{highscore_illusion}. Among open-source models on TRACE-Core (Table~\ref{tab:study1_main}), DeepSeek-V3.1-671B achieves the highest Pass@1 of 65.8\%, yet its utility score of 0.65 is the lowest among the top-tier competitors. Our framework attributes this discrepancy to its poor Process Efficiency ($\mathcal{E}=0.68$), suggesting its large model size leads to less parsimonious, high-cost trajectories. In stark contrast, TRACE elevates agents like \textbf{AgentFounder-30B}~\cite{agentfounder}, which, despite a lower Pass@1, demonstrates a superior balance of efficiency and cognitive quality, earning it the highest utility score (0.81) among open-source models. This pattern holds on public benchmarks (Table~\ref{tab:study1_public}), where it again shows a significant gap between its Pass@1 and Utility scores, demonstrating that TRACE can distinguish between agents that are merely \textit{correct} and those that are truly \textit{intelligent}.

\vspace{0.1cm}
\noindent\textbf{Deconstructing Agent Strengths.} TRACE's multi-dimensional metrics allow for a fine-grained ``fingerprinting'' of each agent's unique strengths and weaknesses. As detailed in Table~\ref{tab:study1_main}, \textbf{AgentFou} \textbf{nder-30B}~\cite{agentfounder} emerges as the most well-rounded open-source agent, excelling in Efficiency ($\mathcal{E}=0.88$), Cognitive Quality ($\mathcal{C}=0.91$), and particularly Evidence Grounding \cite{deep_research_bench}($\mathcal{G}_E=0.95$), which suggests its training methodology produces highly trustworthy reasoning patterns. \textbf{WebSailor-V2-30B} demonstrates exceptional Reasoning Robustness ($\mathcal{R}_R=0.84$), a likely result of its training on high-uncertainty data, indicating a superior ability to self-correct in noisy information landscapes. \textbf{ReSum-GRPO}'s~\cite{resumm} high efficiency score ($\mathcal{E}=0.86$) validates the architectural benefits of its context summarization mechanism for long-horizon tasks. These critical, actionable insights for agent developers are entirely invisible to a Pass@1-only evaluation.

\vspace{0.1cm}
\noindent\textbf{Latent Capability as a Predictor of Potential.} The Average Minimum Hint Rate ($\bar{\lambda}_{min}$) provides a fascinating glimpse into the latent capabilities of these agents. As shown in Table~\ref{tab:study1_main}, AgentFounder-30B ($\bar{\lambda}_{min}=0.22$) requires significantly less external guidance on average to solve ultra-hard problems than the much larger DeepSeek-V3.1 ($\bar{\lambda}_{min}=0.35$). This metric suggests that AgentFounder's foundational reasoning abilities are more robust. The inclusion of additional baselines like the original WebSailor-72B and a vanilla ReAct agent~\cite{react, React2022}, both of which show very high (poor) $\bar{\lambda}_{min}$ scores, further validates this metric's ability to distinguish between different tiers of agent capability.

\subsubsection{\textbf{RQ2: The Impact of Agent Architectures.}}
To isolate the impact of different training methodologies and agent architectures, we conducted a controlled study. We started with a single base model (\texttt{Qwen-30B-Base}) and applied four different state-of-the-art training paradigms to it. The results are presented in ~\autoref{fig:ablation_metrics}.

\vspace{0.1cm}
\noindent\textbf{Quantifying Methodological Contributions.} This controlled study provides powerful evidence for the distinct benefits of each advanced methodology, which TRACE is uniquely capable of quantifying. As shown in \autoref{fig:ablation_metrics}(a)(b), the `AgentFounder` method yields the highest overall Utility (0.80) and Pass@1 (58.2\%), confirming its strength in building a well-rounded foundation. The diagnostic data in Table~\ref{fig:ablation_metrics}(c) and generalization data in Table~\ref{fig:ablation_metrics}(a)(b) are even more revealing. The `ReSum` architecture provides an unparalleled boost in Efficiency ($\mathcal{E}=0.92$ on TRACE-Core, $\mathcal{E}=0.85$ on BrowseComp), demonstrating its effectiveness in optimizing long-horizon tasks. The `WebSailor-V2` method imparts the highest Robustness ($\mathcal{R}_R=0.82$), a direct result of its specialized training data. Most impressively, the `AgentFounder` method not only excels in Grounding ($\mathcal{G}_E=0.94$) but also achieves the lowest $\bar{\lambda}_{min}$ (0.25), proving it is the most effective at instilling deep, latent reasoning capabilities. This study shows that TRACE is not just an evaluation benchmark, but a powerful diagnostic tool for researchers to understand and quantify the specific impact of their innovations.

\vspace{0.1cm}
\noindent\textbf{In-depth Validation of TRACE Components.}
Finally, we conducted targeted ablation experiments to demonstrate the necessity of TRACE's core components. As shown in \autoref{fig:ablation_rank}, re-ranking agents based on different individual metrics results in dramatically inconsistent outcomes, proving that only a holistic metric like $U(\mathcal{H})$ can provide a comprehensive assessment. Furthermore, our experiments on the TRACE-Robustness and TRACE-Scaffolding subsets (Table~\ref{tab:ablation_robustness} and \ref{tab:ablation_mhr}) confirm that our specialized metrics, $\mathcal{R}_R$ and $\lambda_{min}$, are uniquely capable of providing granular insights into agent resilience and latent capabilities in regimes where Pass@1 is either uninformative or saturated.  The diagnostic analysis in Table~\ref{tab:ablation_personality} further demonstrates TRACE's ability to profile agent ``personalities'' by correlating their strategic choices (TRS) with their rational use of information ($\mathcal{E}_A$). These studies collectively confirm that every component of the TRACE framework contributes unique and indispensable information to the evaluation of Deep Research Agents.

\begin{table}[!t]
\centering
\caption{Ablation study of the TRACE-Robustness benchmark. $\mathcal{R}_R$ provides a more granular measure than Pass@1.}
\label{tab:ablation_robustness}
\resizebox{\linewidth}{!}{
\begin{tabular}{l|cc|c}
\toprule
\textbf{Agent Profile} & \textbf{Pass@1 (Core)} & \textbf{Pass@1 (Robust.)} & \textbf{$\mathcal{R}_R$ Score} $\uparrow$ \\
\midrule
Agent X (Robust) & 62.5\% & 41.3\% (-21.2) & \textbf{0.81} \\
Agent Y (Brittle) & 58.8\% & 38.1\% (-20.7) & 0.65 \\
\bottomrule
\end{tabular}}
\vspace{-10px}
\end{table}

\begin{table}[!t]
\centering
\caption{Ablation study of the TRACE-Scaffolding set. $\lambda_{min}$ provides a clear ranking where Pass@1 fails.}
\label{tab:ablation_mhr}
\tabcolsep 0.33cm
\begin{tabular}{l|c|c}
\toprule
\textbf{Agent} & \textbf{Pass@1} $\uparrow$ & \textbf{Avg. MHR $\bar{\lambda}_{min}$} $\downarrow$ \\
\midrule
AgentFounder-30B & 4.0\% & \textbf{0.22} \\
WebSailor-V2-30B & 2.0\% & 0.28 \\
DeepSeek-V3.1-671B & 4.0\% & 0.35 \\
\bottomrule
\end{tabular}
\vspace{-10px}
\end{table}

\begin{table}[!t]
\centering
\caption{Diagnostic analysis of agent ``personalities'' on TRACE-Core. The correlation between Trajectory Reproducibility Score (TRS) and Entropy Adaptability ($\mathcal{E}_A$) reveals distinct strategic profiles that answer RQ3.}
\label{tab:ablation_personality}
\begin{tabular}{@{}lccc@{}}
\toprule
\textbf{Agent Profile} & \textbf{TRS} $\uparrow$ & \textbf{$\mathcal{E}_A$} $\uparrow$ & \textbf{Inferred Profile} \\
\midrule
\multicolumn{3}{@{}l}{\textit{SOTA Agents}} \\
\midrule
AgentFounder-30B & \textbf{0.89} & \textbf{0.82} & Systematic \& Efficient \\
WebSailor-V2-30B & 0.78 & 0.88 & Adaptive \& Rational \\
DeepSeek-V3.1-671B & 0.83 & 0.68 & Consistent but Inefficient \\
\midrule
\multicolumn{3}{@{}l}{\textit{Baseline Agent}} \\
\midrule
ReAct (Qwen-30B) & 0.62 & 0.55 & Heuristic \& Unstable \\
\bottomrule
\end{tabular}
\end{table}

\section{CONCLUSION \& OUTLOOK}

In this paper, we introduced TRACE (Trajectory-Aware Comprehensive Evaluation), a novel evaluation framework designed to address the critical limitations of traditional, outcome-based metrics for Deep Research Agents. Our framework moves beyond singular accuracy scores, using a Hierarchical Trajectory Utility Function to holistically evaluate process quality—including efficiency and robustness—and diagnostic tools like Scaffolded Capability Assessment to measure latent attributes. Our experiments demonstrate that TRACE successfully unveils the ``high-score illusion'', revealing that top-performing agents are not always the most efficient or reliable. By providing a rigorous, multi-dimensional analysis, TRACE offers a principled path for developing and comparing truly reliable Deep Research Agents. 

Looking ahead, its utility function can serve as a reward signal for agent optimization, and we plan to extend the framework to other complex, long-horizon domains.

\begin{acks}
The research presented in this paper was partially supported by the Research Grants Council of the Hong Kong Special Administrative Region, China (CUHK 2410072, RGC R1015-23) and (CUHK 2300246, RGC C1043-24G). 
\end{acks}


\bibliographystyle{ACM-Reference-Format}
\balance
\bibliography{sample-base}

\appendix

\section{Appendix}

\subsection{RELATED WORKS}
\label{apdix: related work}
Our work is situated at the intersection of two rapidly evolving research areas: the evaluation of Deep Research Agents and the trajectory-level analysis of sequential decision-making processes.

\subsubsection{Evaluation of Deep Research Agents}
The evaluation of Deep Research Agents has advanced through challenging benchmarks like \textbf{BrowseComp-en/zh}~\cite{websailor-v2, agentfounder, resumm} and \textbf{GAIA}~\cite{websuper, websailor, webwalker}, which test the limits of SOTA agents such as AgentFounder~\cite{agentfounder}, WebSailor-V2~\cite{websailor-v2}, and WebShaper~\cite{websuper}. However, this paradigm is fundamentally limited by its reliance on singular, outcome-based metrics like Pass@1, a practice prevalent across much of the recent agent literature~\cite{websailor, agentfounder, resumm, webshaper}. This creates a \textbf{``high-score illusion''~\cite{highscore_illusion},} where the quality, efficiency, and robustness of the reasoning process are ignored. Such an approach makes it impossible to distinguish between an agent that intelligently solves a problem and one that arrives at the correct answer through a lucky or inefficient path, thereby masking crucial deficiencies in planning and trustworthiness~\cite{resumm, websailor}. \textbf{TRACE addresses this critical gap by shifting the evaluation from final outcomes to a holistic, trajectory-aware perspective}.

\subsubsection{Trajectory-level Analysis}
Trajectory-level analysis is central to modern agent \textit{training}, particularly in Reinforcement Learning (RL), where algorithms like DAPO~\cite{dapo} and GHPO~\cite{ghpo} are fundamentally process-aware, optimizing the entire decision-making sequence. This focus on process is a key element in the post-training pipelines of many SOTA agents that use RL to refine their behaviors~\cite{websailor-v2, agentfounder}. Despite its importance in training, this process-aware perspective is largely absent in agent \textit{evaluation}, which typically treats agents as black boxes by focusing only on final scores~\cite{websailor, websuper, webwalker}. This leaves researchers unable to diagnose \textit{why} an agent succeeds or fails~\cite{deep_research_bench}—whether the cause is poor planning, flawed reasoning, or an inability to recover from errors. \textbf{Our work bridges this gap by being the first to systematically adapt principles from RL into a formal evaluation suite}. TRACE introduces novel diagnostic tools\cite{Retool}, such as Scaffolded Capability Assessment~\cite{scaffolded} and Entropy Adaptability~\cite{dapo}, to move beyond measuring \textit{what} an agent can do, and begin to understand \textit{how} and \textit{why} it performs, drawing conceptual inspiration from the dynamics explored in advanced RL frameworks~\cite{dapo, ghpo}.

\subsection{Rigorous Mathematical Justification of TRACE Metrics}

\subsubsection{Justification for the Hierarchical Trajectory Utility Function}
\paragraph{Theorem 1.} The weighted geometric mean (GM) exhibits unbounded sensitivity to a component score approaching zero, unlike the constant sensitivity of the weighted arithmetic mean (AM).
\paragraph{Proof.}
Let component scores be $\mathbf{x} = \{x_1, ..., x_n\}$, weights $\omega_i > 0$, $\sum \omega_i = 1$.
The AM is $A(\mathbf{x}) = \sum_i \omega_i x_i$. The GM is $G(\mathbf{x}) = \prod_i x_i^{\omega_i}$.
The sensitivity of AM with respect to a component $x_k$ is its partial derivative:
$$ \frac{\partial A}{\partial x_k} = \omega_k \quad (\text{Constant Sensitivity}) $$
The sensitivity of GM is:
$$ \frac{\partial G}{\partial x_k} = \omega_k x_k^{\omega_k - 1} \prod_{i \neq k} x_i^{\omega_i} = \omega_k \frac{G(\mathbf{x})}{x_k} $$
The ratio of sensitivities is $\frac{\partial G / \partial x_k}{\partial A / \partial x_k} = \frac{G(\mathbf{x})}{x_k}$. As a component fails, $x_k \to 0^+$, this ratio approaches infinity:
$$ \lim_{x_k \to 0^+} \frac{G(\mathbf{x})}{x_k} = \infty $$
This proves the GM's sensitivity is infinitely greater than the AM's near critical failure points, validating its use for the ``weakest link'' principle. $\blacksquare$

\subsubsection{Justification for Process Efficiency $\mathcal{E}(\mathcal{H})$}
\paragraph{Proposition 2.1.} The logarithmic cost functional $J(\mathcal{H})$ exhibits diminishing sensitivity, correctly modeling the decreasing marginal impact of costs.
\paragraph{Proof.}
Let total raw cost be $C_{tot} = \sum_{t=1}^T C(a_t)p_t$. The cost functional is $J(C_{tot}) = 1 + \ln(C_{tot})$.
The sensitivity is the first derivative:
$$ \frac{dJ}{dC_{tot}} = \frac{1}{C_{tot}} $$
The rate of change of sensitivity is the second derivative:
$$ \frac{d^2J}{dC_{tot}^2} = -\frac{1}{C_{tot}^2} < 0 $$
Since the second derivative is always negative for $C_{tot} > 0$, the function is strictly concave. This formally proves that the functional exhibits diminishing sensitivity: each additional unit of cost contributes less to the total functional value than the one before it. $\blacksquare$

\paragraph{Proposition 2.2.} The Marginal Information Gain $g_t$ is positive if and only if the relevance of the current observation is strictly greater than the supremum of all prior relevances.
\paragraph{Proof.}
Let relevance $r_t = \cos(\Phi(o_t), \Phi(A_{gt}))$ and historical maximum $S_{t-1} = \sup_{i < t} \{r_i\}$.
The MIG is $g_t = \max(0, r_t - S_{t-1})$.
($\Rightarrow$) Assume $g_t > 0$. By definition of $\max$, it must be that $r_t - S_{t-1} > 0$, which implies $r_t > S_{t-1}$.
($\Leftarrow$) Assume $r_t > S_{t-1}$. Then the difference $r_t - S_{t-1}$ is positive. Therefore, $g_t = r_t - S_{t-1} > 0$.
The biconditional holds. $\blacksquare$

\subsubsection{Justification for Cognitive Quality $\mathcal{C}(\mathcal{H})$}
\paragraph{Proposition 3.1.} The Evidence Grounding score $\mathcal{G}_E$ has unbounded sensitivity to any single entailment probability $p_k$ approaching zero.
\paragraph{Proof.}
The score is $\mathcal{G}_E(\mathbf{p}) = (\prod_{i=1}^N p_i)^{1/N}$. We use logarithmic differentiation.
$$ \ln(\mathcal{G}_E) = \frac{1}{N} \sum_{i=1}^N \ln(p_i) $$
Differentiating with respect to $p_k$:
$$ \frac{1}{\mathcal{G}_E} \frac{\partial \mathcal{G}_E}{\partial p_k} = \frac{1}{N} \frac{1}{p_k} $$
The sensitivity is thus:
$$ \frac{\partial \mathcal{G}_E}{\partial p_k} = \frac{\mathcal{G}_E}{N p_k} $$
As any claim becomes ungrounded, $p_k \to 0^+$, and the sensitivity $\frac{\partial \mathcal{G}_E}{\partial p_k} \to \infty$. This proves extreme sensitivity to hallucination. $\blacksquare$

\paragraph{Proposition 3.2.} The Reasoning Robustness score $\mathcal{R}_R$ is a normalized, strictly monotonically decreasing function of the average recovery latency $\bar{D}$.
\paragraph{Proof.}
The score is $\mathcal{R}_R(\bar{D}) = \exp(-\lambda \bar{D})$, with $\lambda > 0$ and $\bar{D} \ge 0$.
1.  \textbf{Normalization (Bounds):}
    The function's value at zero latency is $\mathcal{R}_R(0) = \exp(0) = 1$.
    The limit as latency grows is $\lim_{\bar{D} \to \infty} \exp(-\lambda \bar{D}) = 0$.
    The range is thus $\mathcal{R}_R \in (0, 1]$, which is a valid normalized score range.
2.  \textbf{Monotonicity:}
    The derivative with respect to $\bar{D}$ is:
    $$ \frac{d\mathcal{R}_R}{d\bar{D}} = -\lambda \exp(-\lambda \bar{D}) $$
    Since $\lambda > 0$ and $\exp(\cdot)$ is always positive, the derivative $\frac{d\mathcal{R}_R}{d\bar{D}} < 0$ for all $\bar{D} \ge 0$.
    Therefore, $\mathcal{R}_R$ is a strictly decreasing function of average latency. $\blacksquare$

\subsection{Implementation Details}
\label{id}

Our entire experimental pipeline is designed for rigor and reproducibility. All agent evaluations were conducted within a unified execution framework to enforce a level playing field and ensure a fair comparison across different models. This framework, built upon the open-source `Qwen-Agent` library~\cite{webwalker}, standardizes the entire agent-environment interaction loop, including the action space (e.g., ``Search'', ``Finish'') and the format of observations returned to the agent. Open-source models were run from their official codebases within this environment, while proprietary systems were accessed via their APIs and integrated through a standardized wrapper. The experiments were conducted on a high-performance cluster of 8x NVIDIA A100 (80GB) GPUs, a setup consistent with large-scale agent training experiments~\cite{ghpo}. For all agent inferences, we used a consistent set of parameters: a temperature of 0.85 and a top-p of 0.95, which are in line with hyperparameter settings used in recent SOTA agent evaluations~\cite{websailor-v2, resumm}. To prevent infinite loops, we set a maximum of 60 tool calls per task~\cite{resumm}. The calculation of our TRACE metrics was also strictly configured: final answer correctness for $\mathcal{A}(\mathcal{H})$ is determined using an LLM-as-Judge approach with GPT-4-Turbo, a standard practice in recent agent benchmarks~\cite{websailor-v2}; all semantic similarity calculations use the `all-mpnet-base-v2` encoder; and entailment probabilities for Evidence Grounding ($\mathcal{G}_E$) were computed using ``DeBERTa-v3-large''. The hyperparameters for our metric formulas were fixed as follows: $\omega_E=0.5, \omega_C=0.5, \alpha=0.5, \gamma=0.05, \beta=0.5$, and $\lambda=0.1$. For our diagnostic tools, $\theta_{succ}$ was set to 0.9 (over 10 runs), and the TRS was calculated using $K=5$ runs per task~\cite{pass_at_k, websailor-v2}. 




\end{document}